\documentclass{article}

\usepackage[preprint]{neurips_2025}

\title{Assessing Adaptive World Models in Machines \\ with  \textit{Novel Games}}

\usepackage{graphicx}%
\usepackage{multirow}%
\usepackage{amsmath,amssymb,amsfonts}%
\usepackage{amsthm}%
\usepackage{mathrsfs}%
\usepackage[title]{appendix}%
\usepackage{xcolor}%
\usepackage{textcomp}%
\usepackage{manyfoot}%
\usepackage{booktabs}%
\usepackage{algorithm}%
\usepackage{algorithmicx}%
\usepackage{algpseudocode}%
\usepackage{listings}%
\usepackage[colorlinks=true, citecolor=purple, linkcolor=black, urlcolor=blue]{hyperref}
\usepackage{tabularx}
\usepackage{caption}

\begin{document}









\author{%
Lance Ying$^{1,2}$, \quad Katherine M. Collins$^{1,3}$, \quad Prafull Sharma$^{1}$, \quad Cédric Colas$^{1}$, \\ 
\textbf{Kaiya Ivy Zhao$^{1}$}, \quad \textbf{Adrian Weller}$^{3}$, \quad \textbf{Zenna Tavares}$^{4}$, \quad \textbf{Phillip Isola}$^{1}$, \\ \textbf{Samuel J. Gershman}$^{2}$,  \quad\textbf{Jacob D. Andreas}$^{1}$, \quad\textbf{Thomas L. Griffiths}$^{5}$, \\ 
\textbf{François Chollet}$^{6}$,\quad  \textbf{Kelsey R. Allen}$^{7\ddagger}$, \quad \textbf{Joshua B. Tenenbaum}$^{1\ddagger}$ \\ \\
$^1$MIT \quad $^2$Harvard University \quad $^3$University of Cambridge \quad $^4$Basis Research Institute \\ $^5$ Princeton University \quad 
\quad $^6$ ARC Prize Foundation \quad $^7$ University of British Columbia  \\
$^{\ddagger}$ Co-senior authors \\ \\ Correspondence to \texttt{lanceying@mit.edu}}



\maketitle
\begin{abstract}
Human intelligence exhibits a remarkable capacity for rapid adaptation and effective problem-solving in novel and unfamiliar contexts. We argue that this profound adaptability is fundamentally linked to the efficient construction and refinement of internal representations of the environment, commonly referred to as \textit{world models}, and we refer to this adaptation mechanism as \textbf{\textit{world model induction}}. However, current understanding and evaluation of world models in artificial intelligence (AI) remains narrow, often focusing on static representations learned from training on massive corpora of data, instead of the efficiency and efficacy in learning these representations through interaction and exploration within a novel environment. In this Perspective, we provide a view of world model induction drawing on decades of research in cognitive science on how humans learn and adapt so efficiently; we then call for a new evaluation framework for assessing adaptive world models in AI. Concretely, we propose a new benchmarking paradigm based on suites of carefully designed games with genuine, deep and continually refreshing novelty in the underlying game structures --- we refer to this class of games as \textbf{\textit{novel games}}. We detail key desiderata for constructing these games and propose appropriate metrics to explicitly challenge and evaluate the agent's ability for rapid world model induction. We hope that this new evaluation framework will inspire future evaluation efforts on world models in AI and provide a crucial step towards developing AI systems capable of human-like rapid adaptation and robust generalization --- a critical component of artificial general intelligence.
\end{abstract}

\section{Introduction}\label{sec1}

A hallmark of human intelligence is the capacity for rapid adaptation, solving new problems quickly under novel and unfamiliar conditions. Over evolutionary timescales, this adaptive intelligence has enabled humans to survive and flourish in a vast landscape of complex and ever-changing environments. In modern life, people are continually adapting to new social situations such as new laws, cultural environments, partners and foes---often (if not always) with remarkable effectiveness and efficiency. 

Decades of research in cognitive science suggests that a key mechanism supporting this rapid adaptation is the construction and refinement of mental models and intuitive theories to explain the world \citep{johnson1983mental, gopnik2012reconstructing,gelman2011concepts, tenenbaum2011grow, gerstenberg2017intuitive, ullman2020bayesian}. 

In the field of AI, these internal representations are often referred to as ``world models'' \citep{ha2018world}, an agent's representation of its environment, including objects, agents, and causal structures, which can be used to simulate and reason about the world. Building AI systems with more human-like world models and world-modeling capacities has been hypothesized as a crucial step towards building more general intelligent systems.  
The concept of world models has thus garnered significant recent interest in AI research, particularly regarding their structure, how they can be assessed, and whether today's AI systems truly possess them \citep{zhu2024sora, ding2024understanding, hao2023reasoning, andreas2024worldmodels, vafa2024evaluating}.  

However, despite increasing attention, the current ways that internal models are characterized and evaluated in AI systems often diverge importantly from the ways mental models have been studied in humans. Much existing evaluation focuses on static, low-level domain-specific representations learned from large, pre-collected datasets. In contrast, decades of cognitive science research highlights the ways human world models not only support rapid adaptation but are themselves highly adaptive. Our models are dynamically constructed and rapidly adjusted for new domains through active interaction, not merely learned offline from vast corpora. They operate across multiple scales of space, time and abstraction, with higher-level models constraining inferences and induction at lower levels and lower levels grounding the predictions of higher-level abstractions. In this Perspective, we refer to these capacities broadly as a capacity for \textbf{\textit{world model induction}}, which allows intelligent systems to quickly form and validate hypotheses about how new environments and tasks work, and use these hypotheses to guide action, exploration, and bootstrap further and faster learning.

We expect that building and evaluating AI systems capable of this kind of rapid world model induction will be critical for achieving robust, general AI capable of functioning effectively in the complex and fast-changing real world, and especially in {\em human} worlds -- the environments that human beings have evolved in, created, and are continually changing and re-creating. 

To drive AI progress towards this goal, and to be able to measure that progress, we argue for a comprehensive evaluation framework grounded in the cognitive science theories and experimental paradigms that have been used to study world model induction in humans. We propose that games are a uniquely advantageous domain for evaluating these capabilities in AI, given their inherently rich, often hierarchical structures in concepts and skills and their well-controlled environments. Concretely, we introduce an evaluation paradigm centered around the concept of \textit{novel games}. Within this framework, a \textit{novel game} is defined not simply by unseen instances or parametric variations within a familiar game structure, but by environments with structured novelty, where underlying rules, mechanics, object properties, or objectives are initially unknown, hidden, or dynamically changing. Success requires agents to rapidly infer these latent dynamics and causal structures through active, limited interaction and exploration, effectively performing world model induction on-the-fly.

We hope this Perspective will guide future evaluative work on AI world models and thereby driving progress towards machines that can efficiently learn, generalize, and adapt with human-like flexibility and robustness in complex, dynamic real-world environments.

\begin{figure}
    \centering
    \includegraphics[width=1.0\linewidth]{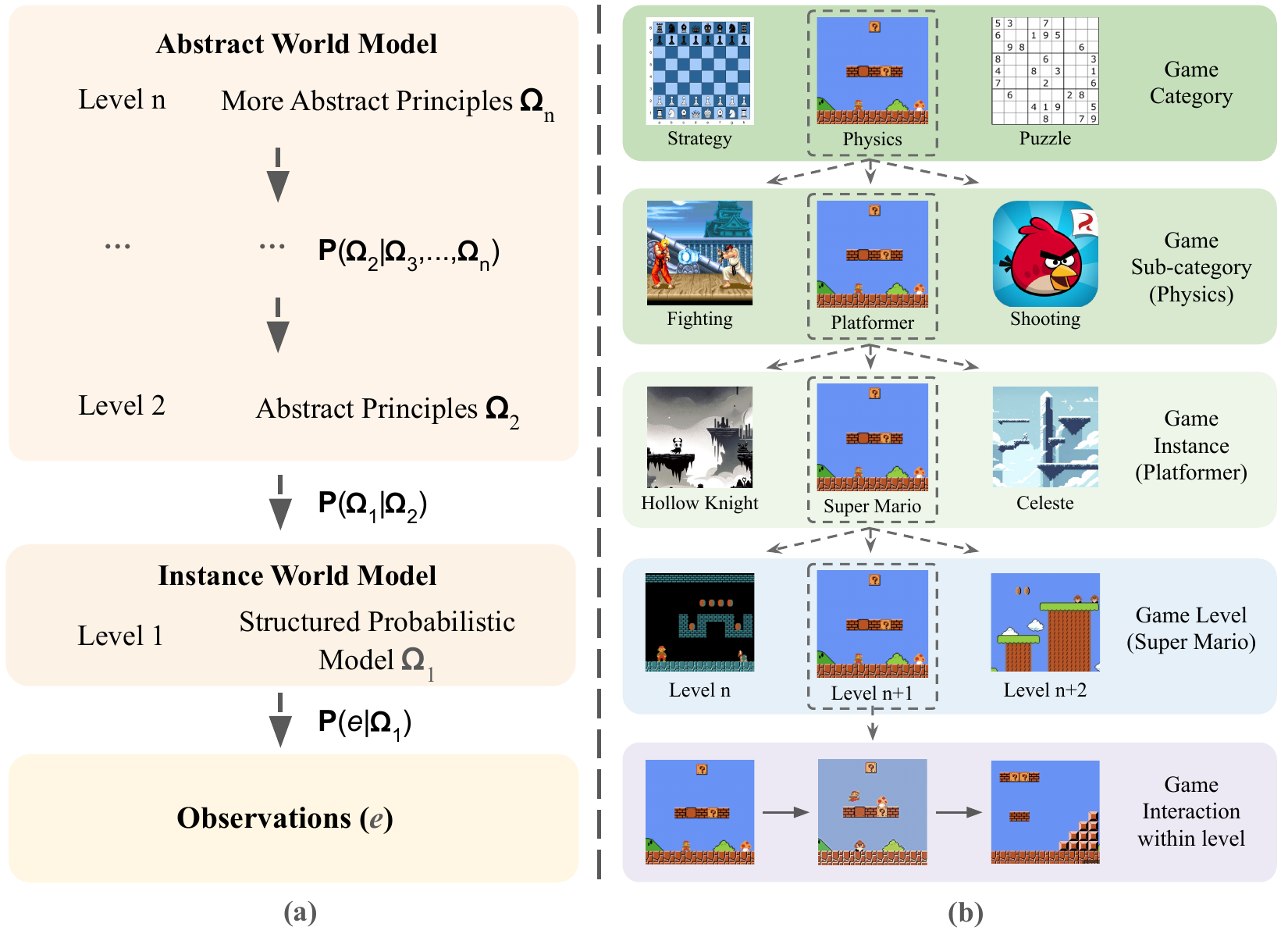}
    \caption{\textbf{Framework for characterizing world models across different levels of abstractions.} \textbf{a)}. World models within a hierarchical Bayesian framework. The structured probabilistic model $\Omega_1$ (ad-hoc world model) generates expectations about possible observations $e$, while abstract knowledge and principles (abstract world model) $\Omega_2, \Omega_3,\dots$ generate the space of possible structures for $\Omega_1$. Each level of abstraction generates hypotheses and probability distributions that support learning at the level below. Then, given observations $e$, the learner can update its world models by inverting the generative model. Figure adapted from \citet{tenenbaum2006theory}. \textbf{b)}. The hierarchical structure of games is analogous to many aspects of the human world model hierarchy. The world model learned at each game level can be ad-hoc and specific. For example, certain game-specific knowledge learned from the first game level (e.g., the effect of consuming a mushroom in Super Mario) can be generalized to the next level (within-domain generalization). On the other hand, meta-learning enables agents to learn domain-general principles at higher levels of abstraction. For example, mastering one platformer game can enable a player to quickly learn to play a different platformer game (cross-domain generalization).}
    \label{fig:world_model}
\end{figure}

\section{World Models in Humans and Machines}\label{sec2}

The precise definition of world models is often debated \citep{ding2024understanding}. For the purposes of this paper, we define \textit{world model} as an agent's internal representation of an environment, including its dynamics, rules, objects, and underlying causal relations. The utility of such a model lies in its ability to allow agents to efficiently simulate different world states for effective decision making, planning, and problem solving. For instance, empirically, compared to model-free reinforcement learning, model-based reinforcement learning has shown to be more data-efficient and affords better generalization \citep{kaiser2019model, moerland2023model}.

The study and evaluation of representations within today's AI systems has garnered significant interest in recent years. However, much of the existing work on AI world models often characterizes and evaluates these representations in a relatively narrow scope, frequently treating them as static representations primarily capturing low-level domain features learned from large, pre-collected datasets, such as whether a model can predict the next frame of a video or future states \citep{lecun2022path, ha2018world}, or recover a visual or spatial representation of the environment \citep{vafa2024evaluating, li2023emergent}. Although these offer valuable insights into the inner representations, we contend that this understanding and evaluation is insufficient for developing AI systems capable of human-level sample efficient learning and adaptation in the open world. 

For adaptation to complex and changing environments, an agent's world models cannot be static representations learned once and fixed. The world is inherently dynamic and often unpredictable. Agents frequently encounter novel situations where previously learned knowledge may be incomplete, only partially applicable, or even become obsolete, necessitating continuous refinement and potentially significant restructuring of the agent's mental model of the environment. Therefore, an adaptive agent's world model must be capable of being dynamically updated and adapted in response to new experience. This continuous process of inferring and revising the world model through interaction is what we refer to as \textbf{world model induction}.

A key characteristic of human world model induction is its \textbf{sample efficiency}. Unlike many current AI paradigms that require training on massive amounts of domain-specific data to learn robust representations, adaptive world model induction necessitates inference of underlying structure and rules from limited, often sparse, online interaction. 

How do humans achieve such remarkable sample efficiency in learning about the world? Extensive research in cognitive science has formulated human learning and intuitive theory induction within a hierarchical Bayesian framework \citep{gopnik2012reconstructing,gelman2011concepts, tenenbaum2011grow,ullman2020bayesian}, where theories and concepts are learned and represented at different levels of abstraction. We can draw a similar conceptual framework for world models, as shown in Figure \ref{fig:world_model}.



In our framework, we partition an agent's world model, denoted as $\Omega$, into two categories of representation, distinguishing between an \textit{instance world model} and \textit{abstract world models}. An instance world model $\Omega_1$, at the lowest level of abstraction, is often a detailed, structured, and domain-specific representation pertaining to a specific instance, which can be constructed on-the-fly to explain observations within that environment, for example, a cognitive map of New York City. Abstract world models $\Omega_2, \Omega_3, \dots$ are more abstract generalizable concepts and principles applicable across domains. For example, one's understanding of real-world physics can be applied even if one moves to a new city. This hierarchical world model structure is key to human adaptation as abstract world models can provide informative priors for a sample-efficient construction of ad-hoc instance world models for interacting and problem-solving within new domains.

Learning can be understood as the construction and refinement of such hierarchical world model. When an agent receives new observations ($e$), its beliefs about the underlying world model $\Omega$ are updated. The posterior probability distribution over possible world models, given new data $e$, is computed by inverting the generative model:
\begin{align}
    P(\Omega | e) & \propto P(e | \Omega) P(\Omega) \nonumber \\ & \propto P(e|\Omega_1)P(\Omega_1|\Omega_2)\dots P(\Omega_n|\Omega_{n+1}) \label{eq:bayesian_update}
\end{align}
Here, the likelihood term $P(e | \Omega)$ represents the probability of observing the data $e$ given a specific world model $\Omega$, while the prior $P(\Omega)$ represents the agent's beliefs about $\Omega$ before the new observation.

The efficacy and sample efficiency of this Bayesian update process are significantly enhanced when the agent is not merely a passive observer but actively seeks out informative data. Insights from cognitive science and developmental psychology, particularly the ``the child as scientist'' framework \citep{schulz2012origins,gopnik1992child,gopnik1996scientist}, suggest that human learning is characterized by hypothesis-driven exploration. Starting at infancy, human learning involves actively planning and designing ``experiments'' --- actions that (intentionally or not) effectively generate informative observations that can discriminate between competing hypotheses about the underlying world model $\Omega$. We expect that this kind of active hypothesis-driven approach to generating data is critical for any intelligent system to converge on an accurate representation of a new, unfamiliar domain with minimal interaction and observation, and more generally for sample-efficient world model induction.

\subsection*{Adaptation via World Model Induction}

Human-like adaptive intelligence necessitates world model induction at different levels of abstraction. This affords agents a number of core behavioral capabilities that are crucial for success in complex and fast-evolving environments. These include:

\begin{enumerate}
    \item \textbf{Rapid Learning in New Domains:} The ability to achieve proficiency quickly in a wide range of previously unseen domains. This learning is facilitated by a combination of mechanisms, including generalization from sparse experiences, efficient goal-directed exploration, and the intelligent use of data sources beyond direct trial-and-error interaction (e.g., information derived from language or social observation).
    
    \item \textbf{Robust Generalization within a Domain:} The capacity to generalize effectively to new and varied situations encountered \textit{within} a newly learned domain. This includes adapting flexibly to novel perceptual inputs, understanding the behavior and affordances of previously unseen object types, handling modified consequences for actions, or pursuing altered goals within that domain's structure.
    
    \item \textbf{Cross-Domain Generalization and Meta-Learning:} The development of meta-learning capabilities, enabling faster and more efficient adaptation to new domains by leveraging prior world models. This reflects the human ability to build generalizable knowledge (e.g., intuitive physics, intuitive psychology) that can bootstrap learning and generalize broadly to new tasks, even those with fundamentally different domain characteristics \citep{spelke2007core, allen2020rapid, lake2017building, chollet2019measure}. However, one also needs to decide which aspects of previous world models are relevant for constructing new world models for a new domain. For example, humans have a general intuitive theory of physics that can broadly be applied to reasoning about physics of objects in any new domain, but humans can also very effectively learn and interact with novel environments that violate basic physical properties \citep{liu2024violations}.
\end{enumerate}

Rapid learning and robust generalization function as complementary components contributing to overall adaptive intelligence and map conceptually onto different facets of the hierarchical Bayesian world model induction formulation. Rapid learning in a new domain focuses on the efficiency of the inference process itself – how quickly the agent can leverage new observations  $e$ to compute and refine the posterior distribution over world models $P(\Omega|e)$ and converge on an accurate world model $\Omega_1$ for a new environment. Robust generalization contributes to a more informative prior $P(\Omega_1 | \Omega_2,...\Omega_n)$ over plausible world models, biasing learning towards likely structures for the new task based on prior experiences and learned abstract principles, from the same (within-domain) or other domains (cross-domain). 

The capacity for rapid world model induction and the associated adaptive capabilities outlined above will be crucial for a wide range of practical applications that AI designers may target, such as adapting to new work environments and collaborating effectively with new human or artificial partners, especially when new tools or protocols are introduced. 
These capacities are also crucial for AI systems intended to function as AI scientists \citep{wang2023scientific,bengio2025superintelligent, ghareeb2025robinmultiagentautomatingscientific, geng2025largelanguagemodelsreliable}, as human scientific discovery fundamentally involves actively forming hypotheses about the world, at different levels of abstraction, and designing experiments to validate these provisional models, mirroring the process of hierarchical world model induction \citep{henderson2010structure}.

Despite the critical importance of world model induction for achieving human-like intelligence, there is a lack of comprehensive evaluation frameworks specifically designed for such world models in AI systems. Existing evaluations of world models in AI models often focus on assessing static world models learned from large, pre-collected datasets or through extensive offline training \citep{vafa2024evaluating, li2023emergent}, rather than measuring the efficiency and flexibility of models in rapidly learning and adapting world models through online exploration and interaction in genuinely novel domains. 

These limitations highlight a critical need for a new evaluation framework designed to comprehensively assess world model in machines. In this perspective, we call for future AI evaluation efforts to holistically assess world models in machines according to the framework outlined above. We contend that games provide particularly rich and controlled environments uniquely well-suited for systematically evaluating rapid model adaptation and the process of world model induction. The inherent structure of many games, with concepts, rules, and skills often organized hierarchically from low-level actions to high-level strategies, aligns well with the requirement for evaluating the learning of world models at different levels of abstraction (See Fig. \ref{fig:world_model}). The remainder of this paper details how games can serve this purpose and proposes a new game-based benchmark proposal in subsequent sections.

\section{Games as a Benchmark for Intelligence}
Games are universal cultural artifacts and have been commonly used as a measure of intelligence \citep{cleveland1907psychology}. While the definition of games is frequently debated, in this paper, we follow previous work on using games to study intelligence \citep{allen2024using} and define games as ``facilitators that structure player behavior and whose main purpose is enjoyment'' \citep{newell1972human}. 

Games have long served as valuable environments for studying machine intelligence by the AI community \citep{campbell2002deep, van2023expertise, silver2016mastering, yannakakis2018artificial, vinyals2019grandmaster, shannon1950, newell1955chess, chase1973mind}. They strike a unique balance by offering clear rules, goals, and feedback while also requiring agents to engage in complex planning, learning, and abstraction. This combination of formal structure and behavioral complexity makes them especially well-suited for probing how intelligent systems\,---\,biological or artificial\,---\,make decisions under uncertainty. Formally, many games can be modeled as Partially Observable Markov Decision Processes (POMDPs), which define a task in terms of hidden states, observations, transitions, and rewards \citep{kaelbling1998planning}. A single game may correspond to one POMDP (e.g., poker), or a distribution over many (e.g., procedurally generated levels in a platformer). This framework has become foundational in both fields for modeling adaptive agents. 



Many established AI benchmarks, particularly those involving complex games  (e.g. Atari \citep{bellemare2013arcade}, Go \citep{silver2016mastering}, StarCraft \citep{vinyals2019grandmaster}) for reinforcement learning agents, follow a training/testing paradigm in which agents are optimized over millions or billions of interaction steps. While such systems can achieve superhuman performance, their success typically reflects extensive optimization within fixed environments rather than rapid, human-like adaptation to new environments. 


To address the limitations of evaluating AI systems solely within the distribution of their training data, various reinforcement learning (RL) benchmarks have introduced forms of task variation to test generalization capabilities. Some approaches involve introducing superficial changes or parametric variations, such as altering initial conditions \citep{plappert2018multi}, perturbing transition dynamics \citep{yu2020meta}, or modifying visual appearance through procedural generation \citep{cobbe2020leveraging}, while crucially keeping the underlying rules and objectives constant. A more ambitious class of benchmarks moves beyond superficial changes by explicitly changing goals or core game mechanics across tasks \citep{plappert2018multi}, or by designing tasks where goals and underlying rules are not directly observable and must be inferred \citep{tsividis2021human, bauer2023human}. While these directions represent valuable steps towards testing broader generalization, they often rely on game structures that are relatively simple (often expressed as a Domain-Specific Language). This simplicity can be a significant limitation, particularly when evaluating the adaptive capabilities of increasingly sophisticated models like Large Language Models (LLMs) or Vision-Language Models (VLMs), which are intended to be deployed in more complex domains.

Most crucially, from the perspective of understanding adaptive intelligence, much generalization evaluation in RL has focused primarily on measuring changes in raw task performance rather than \textbf{evaluating the \textit{process} of adaptation itself}. There is often insufficient focus on whether these tasks truly necessitate the synthesis of new world models, how efficiently an AI model actually constructs these internal representations, or how these models \textit{evolve} over time through interaction with the novel environment.

These limitations collectively highlight a critical gap in current AI evaluation paradigms. While they succeed in measuring performance under varying conditions or with some task generalization, they fall short of assessing the core human capacity to actively construct and dynamically adapt internal world models based on limited online experience and interaction. This ability  demands an evaluation framework specifically tailored to reveal an agent's model-building capabilities. To address this need, we next introduce an evaluation paradigm centered on the use of \textit{novel games} and detail how the design of these games, alongside appropriate metrics, can provide a robust method for assessing rapid world model induction in AI.

\section{Assessing Adaptive World Modeling in AI with \textit{Novel Games}}

In this section, we propose an evaluation paradigm based on a class of games we call \textbf{\textit{novel games}} for assessing the capacity for adaptive world modeling in AI. We are using the phrase \textit{novel games} to refer to \textbf{games with genuine, deep, and continually refreshing novelty in the structure of the environments and goals} presented to the players.  These games require players to construct new world models or modify their models when first learning the game and dynamically throughout their play, across boards, screens or levels.  This contrasts with many existing evaluation framework where AI models are evaluated on domains that are either familiar or highly overtrained, or at most parametric variations of existing domains that the models have been trained on. In the following sections, we first discuss the desiderata for such games, and then we propose a set of metrics for evaluating AI systems within them.

\begin{figure}
    \centering
    \includegraphics[width=1.0 \linewidth]{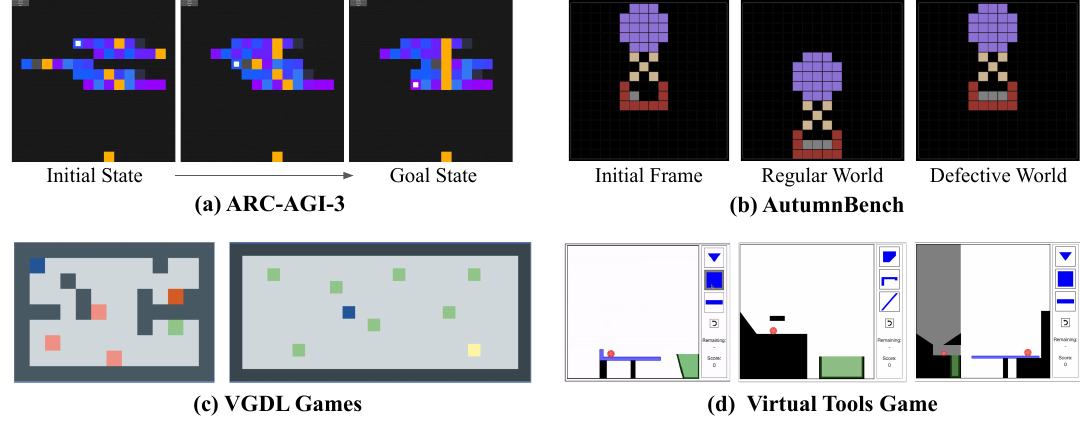}
\captionsetup{labelformat=empty}
    \caption{\quad\quad Figure 2. \textbf{Case studies of novel environments for testing world model induction} \vspace{0.2cm}\\ \textbf{a)} \href{https://arcprize.org/arc-agi}{ARC-AGI-3} \citep{arc2025v3} is an interactive reasoning benchmark. Players are not given instructions about the gameplay or win conditions; instead, they must infer game rules and objectives through interaction. In the featured example, players can move the tiles horizontally. The objective is to align the tiles so that the cells matching the standalone cell at the bottom—the yellow cell, in this case—are collinear. \vspace{0.2cm} \\  \textbf{b)} \href{https://autumn.basis.ai/}{AutumnBench} \citep{basis2025autumn} evaluates an agent's ability to discover latent mechanics through interactive exploration of grid-world environments. The evaluation follows a two-phase protocol: interaction and test. During interaction, agents explore environments freely without rewards or goals. The subsequent test phase evaluates their understanding through three tasks: masked frame prediction, defect detection, and planning. In the featured example, through interaction the agent is expected to discover the rule that adding a grey block lowers the ballon, which makes the right-most frame an anomaly. \vspace{0.2cm}\\ \textbf{c)} \href{https://pedrotsividis.com/vgdl-games/}{VGDL games} \citep{schaul2013video,perez2019general} were used by \citet{tsividis2021human} to evaluate an agent's capability to discover rules and objectives in ambiguous environments when no language instructions are provided. In their games, the agent can press a few keys on a keyboard to explore the environment to pass each level. Humans generally learn to play these new games in a matter of minutes. \vspace{0.2cm} \\ \textbf{d)} \href{https://sites.google.com/view/virtualtoolsgame}{Virtual Tools Game} \citep{allen2020rapid} tests rapid learning in physical reasoning scenarios. The goal is to select and drop one of the tools on the right so that the red ball ends up in the green bin. \citet{allen2020rapid} show that humans quickly solve these tasks by leveraging their intuitive understanding of physics. This showcases robust cross-domain generalization.
 }
    \label{fig: case_studies}
\end{figure}

\subsection{Desiderata for Designing \textit{Novel Games}}

The key requirement for our AI evaluation paradigm is the inherent novelty in the underlying game structure. These games must offer genuinely new adaptation challenges, meaning they are significantly distinct from established and widely studied games (like Chess, Go, or classic Atari titles) and necessitate new world models. This distinctiveness is crucial to prevent AI systems from reusing existing world models to solve the task or exploiting readily available online resources such as wikis and walkthroughs that describe optimal strategies for existing popular games.

However, the space of all possible \textit{novel games} is infinitely large as one can construct new games with any arbitrary mechanics. We propose \textit{novel games} should be grounded in the kinds of diverse, highly dynamic and novel environments encountered by humans, thus providing a testbed for how well an AI system can learn and adapt\,---\,relative to humans\,---\,in all kinds of (game) worlds intuitive to humans, either alone or with humans.

\subsubsection{Desideratum 1: Novelty in Game Structures}

In this section, we articulate key features for the design of domains with genuine novelty in their underlying structures for testing the three aspects of adaptive capabilities listed in Section \ref{sec2}:

\paragraph{Rapid Learning and Theory-driven Exploration} We encourage the design of game environments where the underlying mechanics are not fully transparent or pre-specified to the player. Instead, crucial aspects -- including types of objects, specific rules governing interactions, affordances and properties of objects, or the consequences of actions -- should be \textbf{partially or entirely latent}, requiring the agent to infer them through active gameplay, exploration, and experimentation. We highlight three such game environments from previous studies in Figure \ref{fig: case_studies}.

This design compels the AI agent to function as an active learning system, dynamically constructing an understanding of its environment. This exploratory process should ideally reflect aspects of theory-driven learning \citep{ullman2020bayesian, tsividis2021human}. The agent must be capable of forming hypotheses about latent rules or object behaviors based on observation, strategically planning and executing `experiments' through its actions to test these hypotheses, and refining its internal model based on the observed outcomes. Effectively pursuing this form of active, model-building learning necessitates setting \textit{epistemic goals}---objectives focused on acquiring knowledge and reducing uncertainty about the game's state and mechanics. This implies a form of ``planning to learn,'' where the agent deliberately chooses actions not solely for immediate task progress or reward, but to validate its hypotheses and build a more accurate world model, demonstrating an awareness of its own knowledge gaps.

\paragraph{Robust Generalization within a Domain}
Effective adaptation within a learned novel domain requires robustness to variations and changes occurring within that specific environment's structure. \textit{Novel games} should be designed to feature multiple levels or configurations, introduce new object types with distinct properties, modify existing rules or mechanics, or alter goals and outcomes over the course of interaction. Crucially, these games can also incorporate mechanics that cause the environment, including its rules and dynamics, to \textbf{evolve dynamically over time}, potentially influenced by the player's actions or external events. This dynamic aspect necessitates that the agent continuously monitors the environment, detects changes, and updates its internal world model online. As environmental dynamics or rules shift, previously viable goals might become unreachable, requiring the agent to adapt its objectives. This can mirror real-world scenarios where societal laws, technological capabilities, or even aspects of physical systems are not static. 

\paragraph{Flexible Generalization across Domains}
To evaluate the capacity for cross-domain generalization and meta-learning, the benchmark should include game sets where \textbf{abstract principles or models learned can be productively transferred to a new game}, despite significant differences in surface rules or mechanics. For example, training on games involving various scenarios governed by a consistent set of physics rules (e.g., gravity, momentum) allows an agent to induce an abstract ``intuitive physics'' model. This model can then be transferred to accelerate adaptation in a new game featuring new objects and tasks but operating under similar physical laws, enabling the agent to predict outcomes more effectively from the outset. This design measures the agent's ability to acquire generalizable inductive biases and bootstrap learning in new, yet abstractly related, domains.

\subsubsection{Desideratum 2: Intuitive and Learnable for Human Players}
For \textit{novel games} to stress-test AI models' capability to adapt in the human world, their core mechanics and objectives should be fundamentally \textbf{intuitive and learnable for average human players}. This criterion is essential because a key goal of this evaluation paradigm is to measure human-like adaptation skills. Games that humans find intuitive are likely structured in ways that resonate with fundamental human inductive biases: the inherent cognitive predispositions and learning mechanisms shaped by the cultural and physical environment humans inhabit~\citep{allen2024using, dubey2018investigatinghumanpriorsplaying}. Ensuring human learnability serves practical purposes: it allows for benchmarking AI performance directly against human capabilities, provides a valuable constraint on the complexity and potential arbitrariness of the game generation process, and aligns the evaluation with the broader goal of developing AI that can learn from and collaborate with humans in novel scenarios.


\subsubsection{Desideratum 3: Diversity in World Models and Learning Mechanisms}

To span the diverse array of challenges people---and AI systems in a human-world may face---the benchmark game suite should encompass significant \textbf{diversity} to necessitate different \textit{types of world models} that agents are compelled to induce. For example, while some games may primarily involve learning about spatial relationships or object physics, others can require understanding and modeling other agents in multi-agent games (whether competitive or collaborative with other human or artificial agents). Successfully adapting in multi-agent scenarios often requires agents to develop sophisticated mental models about other agents, inferring and representing their goals, beliefs, intentions, or emotional states (often referred to as Theory of Mind \citep{gopnik1992child}), which constitutes a crucial aspect of human social adaptation.

Furthermore, the benchmark should feature diversity in the \textbf{learning mechanisms} available to the agent. Some games can be designed to offer minimal or no explicit instructions, compelling the agent to induce the world model predominantly through interaction and exploration. Conversely, other games could provide structured linguistic instructions, demonstrations, or tutorials, allowing for the evaluation of how well agents can leverage external, often multimodal, information to accelerate model construction. Incorporating scenarios where information about mechanics or objectives is conveyed implicitly through social means, such as Non-Player Characters (NPCs) that demonstrate actions or use language, can provide a critical way to test learning through observation and social scaffolding like people.


\subsubsection{Combining Desiderata in a Generative Framework}

A central challenge in this evaluation paradigm is the continual provision of games that rigorously satisfy our desiderata. Specifically, the inherent novelty of these games is ephemeral; as AI systems (and indeed, humans) gain experience with their mechanics, the games quickly become familiar, thereby undermining their utility as tests of adaptation to truly novel situations.

\begin{figure}
    \centering
    \includegraphics[width=\linewidth]{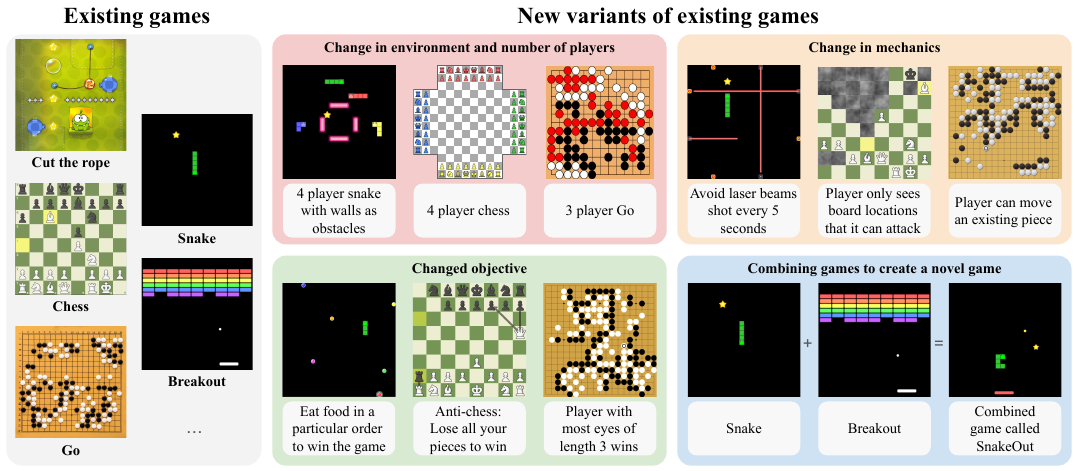}
    \caption{\textbf{Creating \textit{novel games}.} There are many ways that researchers can create \textit{novel games}. Variants could be sourced from existing games by modifying the environment and number of players, mechanics, or objective function. \textit{Novel games} could also be formed by combining existing games.}
    \label{fig:creating-new-games}
\end{figure}

We propose that game benchmarks should be thought of as a \textbf{generative process} over such games which can continually sample new \textit{novel games} that satisfy our desiderata. Like language, games can be compositional and continually reconfigured. Modifications can vary the game mechanics, partners, and other game features (see Figure~\ref{fig:creating-new-games}). This would allow the game benchmark to continue to evolve and cover a large space of novel and diverse environments that AI would need to adapt to, thus mitigating overfitting.

\subsection{Evaluation of AI Agent's World Modeling Capacity}

Once we have designed games that pose meaningful challenges on adaptive world modeling for AI, we need a comprehensive evaluation framework to probe and characterize the internal world models learned by the agent.

\paragraph{Sample Efficiency in Adaptation}

A measure of learning efficiency is how quickly a model can achieve proficiency with limited experience. One could evaluate this by providing a restricted ``budget'' of training attempts (trials) within a game level and assessing performance. This budget can be varied to provide a fine-grained understanding of learning dynamics and adaptability under different constraints. For example, one can measure performance after a fixed number of attempts or trials, or quantify the number of game-plays required to reach average human performance (e.g. \citealt{lake2017building}). 

\begin{figure}
    \centering
    \includegraphics[width=0.9\linewidth]{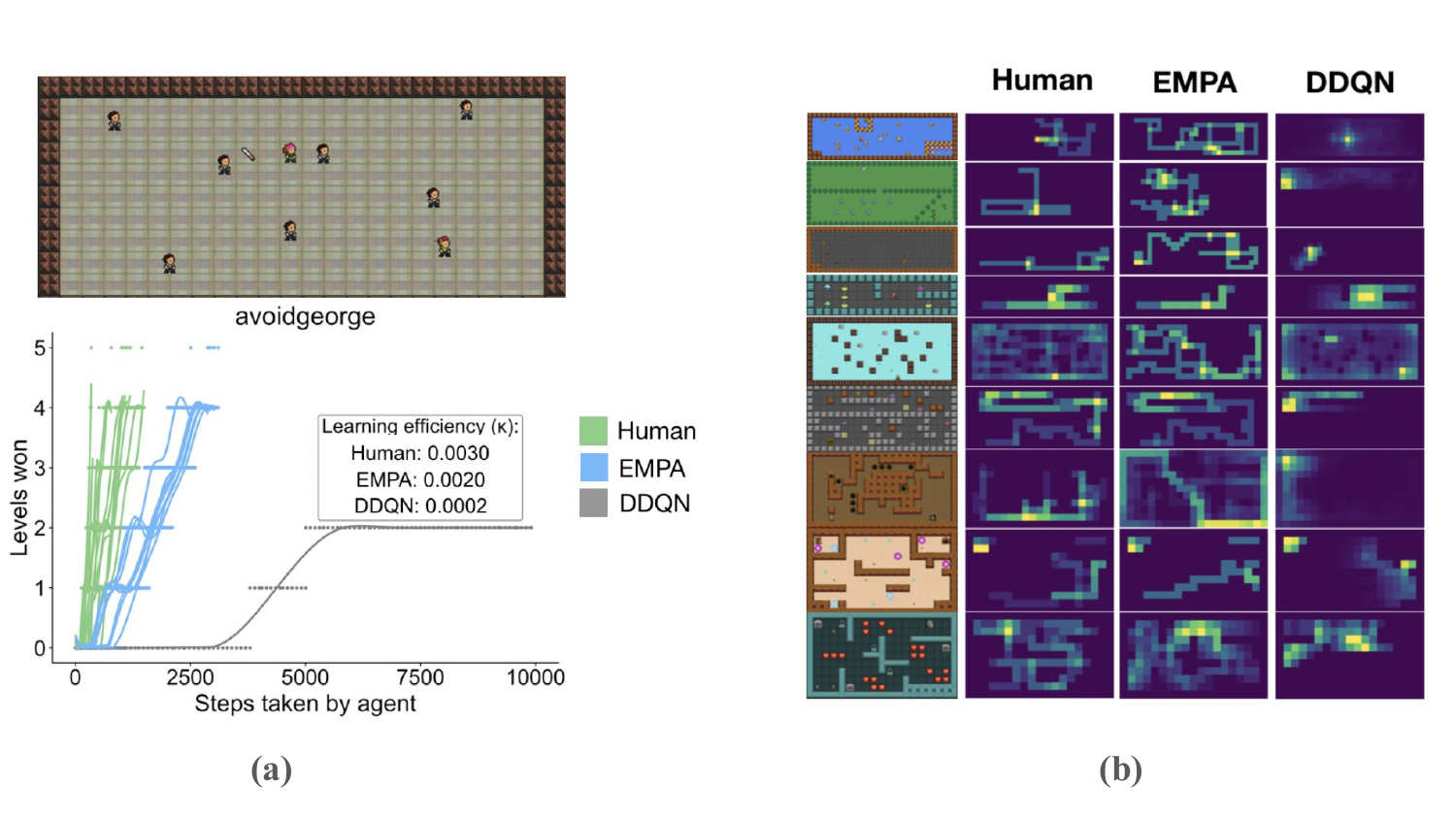}
    \caption{Evaluation methods used by \citet{tsividis2021human} for comparing learning behaviors across humans and models (EMPA and DDQN). \textbf{a)} Quantitative learning efficiencies plotted as the number of levels passed as a function of exploration steps. \textbf{b)} Qualitative analysis of different exploration patterns demonstrated by different learning agents.}
    \label{fig:analysis}
\end{figure}

We feature the work by \citet{tsividis2021human} in Figure \ref{fig:analysis}a as a case study. In this example, they compute learning efficiency scores of different models and plot the success rate of the models at all game levels as a function of learning samples, which allows direct comparison between the sample efficiency of humans vs models at learning and solving new tasks.

\paragraph{Qualitative Analysis of Exploration and Learning Behavior}
Beyond quantitative metrics, a qualitative analysis of how agents explore and learn within novel environments can reveal crucial insights into their capabilities for world model induction. Different learning approaches often manifest in distinct exploration patterns, reflecting their strategies for gathering information and inferring rules. As demonstrated in Figure \ref{fig:analysis}b from \citet{tsividis2021human}, human players tend to exhibit targeted and efficient exploration, focusing on areas relevant to understanding the game mechanics. Similarly, the EMPA model proposed in the same paper shows more structured but sometimes less efficient exploration, while DDQN \citep{van2016deep} often displays highly diffuse and less directed exploration, indicative of a struggle to efficiently form coherent internal representations of the environment. Such qualitative visual comparisons can highlight the differing capacities of agents to induce and utilize effective world models, revealing whether they are genuinely understanding the game's rules or merely optimizing for short-term rewards.



\paragraph{Probing Internal World Models}

To properly assess rapid world model induction, it is essential to gain insight into the nature of the internal representations that the agent constructs and how these representations are dynamically updated in response to new observations and actions.

The methods for inspecting these internal world models are heavily dependent on the agent's architecture. For models based on explicit program synthesis or symbolic reasoning, the inferred world model may be directly interpretable as the synthesized program or set of rules \citep{tsividis2021human, das2023combining}. This offers a transparent view of the agent's current representation of the environment, and this allows direct comparison to the kinds of hierarchical representations in humans.

For neural networks, inspecting the internal world model is more complex, typically involving analyzing representation spaces, activation patterns, attention mechanisms, or tracing reasoning processes through the network \citep{vafa2024evaluating}. Techniques such as probing specific network layers for learned features related to game mechanics or dynamics can reveal aspects of the implicit world model. For many of the large foundation models that can interact with humans in natural language, we can also examine their understanding of the game mechanics through targeted question answering at different levels of abstraction. For instance, one may ask questions about abstract principles, such as knowledge about objects (e.g. object permanence) or properties specific to the game, such as the mass of a particular object.

By probing these internal representations and their changes over time as the agent interacts with a game environment, we can gain crucial qualitative insights into how the AI agent is actively inferring, representing, and revising its understanding of the world. 

\section{How To: Designing a Generative \textit{Novel Game} Benchmark in Practice}

In this section, we offer practical recommendations for how to go about designing game benchmarks under our evaluative paradigm in practice. 

\subsection{Game Design and Creation}

A central practical consideration for our benchmarking paradigm is the generation of a continuous stream of \textit{novel games} required to test adaptive world model induction and generalization across domains. 

One approach is to design every novel game entirely from scratch, which, while offering maximum control over the novelty injected, can be a significantly labor-intensive process, potentially limiting the scale and diversity of the benchmark suite. For instance, ~\citet{zhang2024people} manually created $120$ variants of grid-based games, such as Tic-Tac-Toe and Gomoku. More broadly, drawing inspiration from human creativity, which often builds upon existing structures to create novel variants~\citep{velez2024rise, zhao2024rational, youn2015invention, collins2025gamecreation}, AI benchmark creators can adopt a similar strategy. Instead of solely focusing on generating games disconnected from known forms, a promising avenue involves creating interesting and challenging novel variants by systematically altering the rules, objectives, player counts, or interaction dynamics of existing popular games. This method leverages the existing human familiarity with the base game, providing a potentially more intuitive starting point for both human and AI adaptation to the introduced changes (we highlight some examples of systematic alteration in Fig. \ref{fig:creating-new-games}). 

To overcome the manual effort associated with both designing games from scratch and systematically altering existing ones, benchmark developers can leverage recent advancements in generative AI models. For instance, recent work has explored using LLMs to synthesize entire new game mechanics for benchmarking AI models \citep{verma2025measuring}. By utilizing LLMs, one can formulate sophisticated generative pipelines, potentially employing quality-diversity algorithms \citep{pugh2016quality, cully2017quality, todd2024gavel}. Such methods are well-suited for generating a diverse range of outputs optimized not just for a single objective but across multiple criteria such as variety, difficulty, solvability, and even potential enjoyment \citep{pourcel2023aces, davidson2024goals, zhang2024people, chu2020exploratory}. Yet, while automation can accelerate the process, integrating human feedback or player modeling into the loop remains essential to ensure that the generated games are not only syntactically-valid but also pose meaningful, learnable, and intuitive adaptation challenges for AI. 

\subsection{Human Evaluation and Validation}

Integration of human feedback and participation is indispensable. This is crucial for both validating the suitability of the \textit{novel games} themselves and for providing a vital baseline against which AI performance can be compared.

Human input is essential for ensuring that the \textit{novel games} generated for the benchmark meet the desiderata we lay out above: of being intuitive, adaptable, and learnable for people. Benchmark creators can employ a combination of self-reported data and behavioral measures as proxies for these qualities. For instance, soliciting ratings from human players regarding game engagement, playability, and perceived difficulty can provide valuable qualitative feedback. Additionally, analyzing the playing traces and learning curves of human participants can offer quantitative insights into their ability to infer rules, develop strategies, and adapt to the novel mechanics, thereby confirming that the games are indeed human-learnable and intuitive as intended. For instance, ARC-AGI conducts human evaluations to make sure benchmark tasks can be solved by most human players \citep{chollet2025arc}. In addition to controlled human experiments, one could envision the creation of dedicated game platforms where human players can engage at their convenience or where human players can interact with or compete against AI models, similar to existing initiatives like TextArena~\citep{guertler2025textarena}. This approach offers the advantage of collecting large-scale data in a relatively naturalistic setting. Such human data can be useful for benchmarking AI on these \textit{novel games} and also allow researchers to perform detailed analyses to investigate and compare the process of world model induction in AI systems versus humans.

\section{Discussion and Looking Forward}

In this Perspective, we have argued that a critical component for developing truly general and robust artificial intelligence lies in its capacity for adaptation to novel circumstances. This adaptive capability is fundamentally linked to the agent's ability to rapidly induce and dynamically refine internal world models when confronted with unknown environments. We then introduce an evaluation paradigm centered around carefully constructed \textit{novel games}. This framework is specifically designed to evaluate AI systems on their capacity for adaptive world modeling, which is essential for efficient learning and robust generalization in dynamic, unforeseen environments where underlying rules and structures are often hidden from the agent.

While we believe this paradigm offers a valuable path forward, we acknowledge several important outstanding questions and challenges that warrant future investigation and refinement. A fundamental question that may arise regarding our central thesis is the extent to which hierarchical and adaptive world models are truly necessary for rapid and efficient adaptation. Our Perspective is strongly motivated by and aligned with extensive evidence from cognitive science and developmental psychology \citep{tenenbaum2011grow, gopnik2012reconstructing}, which highlights the crucial role of constructing and refining internal models in human learning and adaptation. However, one may argue that alternative computational mechanisms could facilitate adaptation without relying on world models at all \citep{brooks1991intelligence}. For example, one may question if a reactive or purely end-to-end system with massive capacity might exhibit surprising levels of adaptation through pattern generalization across diverse training data, even if it lacks a structured, predictive model of environmental dynamics. This view has met with some skepticism by recent work in AI, which argues that world models are essential for building general agents \citep{richens2025general}. Nevertheless, the extent to which such model-free approaches can achieve human-level rapid, sample-efficient adaptation and robust generalization remains an open empirical question.

Second, while we can evaluate performance on the tasks within \textit{novel games}, directly measuring the quality and efficiency of internal world model induction at different levels of hierarchy presents its own challenges. Developing metrics that specifically quantify how well an agent has inferred the latent rules or dynamics—beyond just task success—and how efficiently it updates this understanding over time is crucial. This may involve developing probing techniques, counterfactual evaluation methods based on the inferred model, or analyzing the structure and dynamics of internal representations. Advancing the methodology for inspecting and quantifying hierarchical internal representations will be vital for a complete understanding of adaptive world models in machines.

Lastly, there are important open questions about the external validity of findings from novel game benchmarks. While games provide a valuable, controlled, and complex environment for studying aspects of adaptive intelligence and world model induction, it remains an open question how directly the adaptive capabilities demonstrated within these simulated environments will transfer to the complexities, continuous nature, and richness of real-world adaptation challenges. Rigorously assessing and quantifying the external validity of benchmark results --- for instance, by designing complementary evaluations in limited real-world scenarios or high-fidelity, open-ended simulation environments grounded in physical or social realism --- is a critical direction for future research. Such work is necessary to ensure that progress on game benchmarks translates meaningfully to progress on real-world AI capabilities, particularly if AI systems are designed to engage with people~\citep{collins2024building}.


\section{Conclusion}
In this paper, we have argued that building human-like adaptability in machines necessitates adaptive world models, which affords sample efficient world model induction in any new domains. We then proposed a novel framework for assessing adaptive world models centered on the concept of \textit{novel games}. We believe this proposed evaluation paradigm holds significant potential to serve as a core component in assessing current AI models and drive research towards systems that exhibit the rapid, flexible, and robust adaptability characteristic of human intelligence, thus contributing meaningfully to the ambitious pursuit of artificial general intelligence.

\section*{Acknowledgments}

We thank Ced Zhang, Lio Wong, Graham Todd, Mau Barba, Gianluca Bencomo, and Tyler Brooke-Wilson for valuable conversations that informed this work. We thank Greg Kamradt for helpful feedback on the draft. This work is funded in part by AFOSR, ONR through the Science of AI, and MURI programs, a Schmidt AI2050 Fellowship to JBT, and the Siegel Family Quest for Intelligence at MIT. KMC acknowledges support from the Cambridge Trust and King's College Cambridge. AW  acknowledges  support  from  a  Turing  AI  Fellowship  under grant  EP/V025279/1 and the Leverhulme Trust via CFI. This work is supported (in part) by ELSA - European Lighthouse on Secure and Safe AI funded by the European Union under grant agreement No. 101070617 and by the European Union’s Horizon 2020 research and innovation programme under the Marie Skłodowska-Curie grant agreement No 101065949 (CC). JA acknowledges support from the Sloan Foundation and the U.S.\ National Science Foundation under grant IIS-2212310.  Views and opinions expressed are however those of the author(s) only and do not necessarily reflect those of the European Union or European Commission.

\bibliography{sn-bibliography}

\begin{thebibliography}{}

\bibitem[Allen et~al., 2024]{allen2024using}
Allen, K., Br{\"a}ndle, F., Botvinick, M., Fan, J.~E., Gershman, S.~J., Gopnik, A., Griffiths, T.~L., Hartshorne, J.~K., Hauser, T.~U., Ho, M.~K., et~al. (2024).
\newblock Using games to understand the mind.
\newblock {\em Nature Human Behaviour}, pages 1--9.

\bibitem[Allen et~al., 2020]{allen2020rapid}
Allen, K.~R., Smith, K.~A., and Tenenbaum, J.~B. (2020).
\newblock Rapid trial-and-error learning with simulation supports flexible tool use and physical reasoning.
\newblock {\em Proceedings of the National Academy of Sciences}, 117(47):29302--29310.

\bibitem[Andreas, 2024]{andreas2024worldmodels}
Andreas, J. (2024).
\newblock Language models, world models, and human model-building.

\bibitem[{{ARC Prize}}, 2025]{arc2025v3}
{{ARC Prize}} (2025).
\newblock Arc-agi-3.
\newblock Retrieved from https://arcprize.org/arc-agi/3.

\bibitem[Basis, 2025]{basis2025autumn}
Basis (2025).
\newblock Autumnbench: World model learning in humans and ai.

\bibitem[Bauer et~al., 2023]{bauer2023human}
Bauer, J., Baumli, K., Behbahani, F., Bhoopchand, A., Bradley-Schmieg, N., Chang, M., Clay, N., Collister, A., Dasagi, V., Gonzalez, L., et~al. (2023).
\newblock Human-timescale adaptation in an open-ended task space.
\newblock In {\em International Conference on Machine Learning}, pages 1887--1935. PMLR.

\bibitem[Bellemare et~al., 2013]{bellemare2013arcade}
Bellemare, M.~G., Naddaf, Y., Veness, J., and Bowling, M. (2013).
\newblock The arcade learning environment: An evaluation platform for general agents.
\newblock {\em Journal of artificial intelligence research}, 47:253--279.

\bibitem[Bengio et~al., 2025]{bengio2025superintelligent}
Bengio, Y., Cohen, M., Fornasiere, D., Ghosn, J., Greiner, P., MacDermott, M., Mindermann, S., Oberman, A., Richardson, J., Richardson, O., et~al. (2025).
\newblock Superintelligent agents pose catastrophic risks: Can scientist ai offer a safer path?
\newblock {\em arXiv preprint arXiv:2502.15657}.

\bibitem[Brooks, 1991]{brooks1991intelligence}
Brooks, R.~A. (1991).
\newblock Intelligence without representation.
\newblock {\em Artificial intelligence}, 47(1-3):139--159.

\bibitem[Campbell et~al., 2002]{campbell2002deep}
Campbell, M., Hoane~Jr, A.~J., and Hsu, F.-h. (2002).
\newblock Deep blue.
\newblock {\em Artificial intelligence}, 134(1-2):57--83.

\bibitem[Chase and Simon, 1973]{chase1973mind}
Chase, W.~G. and Simon, H.~A. (1973).
\newblock The mind's eye in chess.
\newblock In {\em Visual information processing}, pages 215--281. Elsevier.

\bibitem[Chollet, 2019]{chollet2019measure}
Chollet, F. (2019).
\newblock On the measure of intelligence.
\newblock {\em arXiv preprint arXiv:1911.01547}.

\bibitem[Chollet et~al., 2025]{chollet2025arc}
Chollet, F., Knoop, M., Kamradt, G., Landers, B., and Pinkard, H. (2025).
\newblock Arc-agi-2: A new challenge for frontier ai reasoning systems.
\newblock {\em arXiv preprint arXiv:2505.11831}.

\bibitem[Chu and Schulz, 2020]{chu2020exploratory}
Chu, J. and Schulz, L. (2020).
\newblock Exploratory play, rational action, and efficient search.
\newblock In {\em Proceedings of the Annual Meeting of the Cognitive Science Society}, volume~42.

\bibitem[Cleveland, 1907]{cleveland1907psychology}
Cleveland, A.~A. (1907).
\newblock The psychology of chess and of learning to play it.
\newblock {\em The American Journal of Psychology}, 18(3):269--308.

\bibitem[Cobbe et~al., 2020]{cobbe2020leveraging}
Cobbe, K., Hesse, C., Hilton, J., and Schulman, J. (2020).
\newblock Leveraging procedural generation to benchmark reinforcement learning.
\newblock In {\em International conference on machine learning}, pages 2048--2056. PMLR.

\bibitem[Collins et~al., 2024]{collins2024building}
Collins, K.~M., Sucholutsky, I., Bhatt, U., Chandra, K., Wong, L., Lee, M., Zhang, C.~E., Zhi-Xuan, T., Ho, M., Mansinghka, V., et~al. (2024).
\newblock Building machines that learn and think with people.
\newblock {\em Nature human behaviour}, 8(10):1851--1863.

\bibitem[Collins et~al., 2025]{collins2025gamecreation}
Collins, K.~M., Todd, G., Wong, L., Zhang, C., Togelius, J., Weller, A., Chu, J., Griffiths, T., and Tenenbaum, J. (2025).
\newblock Generation and evaluation in the human invention process through the lens of game design.
\newblock In {\em Proceedings of the Annual Meeting of the Cognitive Science Society}, volume~47.

\bibitem[Cully and Demiris, 2017]{cully2017quality}
Cully, A. and Demiris, Y. (2017).
\newblock Quality and diversity optimization: A unifying modular framework.
\newblock {\em IEEE Transactions on Evolutionary Computation}, 22(2):245--259.

\bibitem[Das et~al., 2023]{das2023combining}
Das, R., Tenenbaum, J.~B., Solar-Lezama, A., and Tavares, Z. (2023).
\newblock Combining functional and automata synthesis to discover causal reactive programs.
\newblock {\em Proceedings of the ACM on Programming Languages}, 7(POPL):1628--1658.

\bibitem[Davidson et~al., 2024]{davidson2024goals}
Davidson, G., Todd, G., Togelius, J., Gureckis, T.~M., and Lake, B.~M. (2024).
\newblock Goals as reward-producing programs.
\newblock {\em arXiv preprint arXiv:2405.13242}.

\bibitem[Ding et~al., 2024]{ding2024understanding}
Ding, J., Zhang, Y., Shang, Y., Zhang, Y., Zong, Z., Feng, J., Yuan, Y., Su, H., Li, N., Sukiennik, N., et~al. (2024).
\newblock Understanding world or predicting future? a comprehensive survey of world models.
\newblock {\em ACM Computing Surveys}.

\bibitem[Dubey et~al., 2018]{dubey2018investigatinghumanpriorsplaying}
Dubey, R., Agrawal, P., Pathak, D., Griffiths, T.~L., and Efros, A.~A. (2018).
\newblock Investigating human priors for playing video games.

\bibitem[Gelman and Legare, 2011]{gelman2011concepts}
Gelman, S.~A. and Legare, C.~H. (2011).
\newblock Concepts and folk theories.
\newblock {\em Annual review of anthropology}, 40(1):379--398.

\bibitem[Geng et~al., 2025]{geng2025largelanguagemodelsreliable}
Geng, J., Chen, H., Arumugam, D., and Griffiths, T.~L. (2025).
\newblock Are large language models reliable ai scientists? assessing reverse-engineering of black-box systems.

\bibitem[Gerstenberg and Tenenbaum, 2017]{gerstenberg2017intuitive}
Gerstenberg, T. and Tenenbaum, J.~B. (2017).
\newblock Intuitive theories.

\bibitem[Ghareeb et~al., 2025]{ghareeb2025robinmultiagentautomatingscientific}
Ghareeb, A.~E., Chang, B., Mitchener, L., Yiu, A., Szostkiewicz, C.~J., Laurent, J.~M., Razzak, M.~T., White, A.~D., Hinks, M.~M., and Rodriques, S.~G. (2025).
\newblock Robin: A multi-agent system for automating scientific discovery.

\bibitem[Gopnik, 1996]{gopnik1996scientist}
Gopnik, A. (1996).
\newblock The scientist as child.
\newblock {\em Philosophy of science}, 63(4):485--514.

\bibitem[Gopnik and Wellman, 1992]{gopnik1992child}
Gopnik, A. and Wellman, H.~M. (1992).
\newblock Why the child's theory of mind really is a theory.

\bibitem[Gopnik and Wellman, 2012]{gopnik2012reconstructing}
Gopnik, A. and Wellman, H.~M. (2012).
\newblock Reconstructing constructivism: causal models, bayesian learning mechanisms, and the theory theory.
\newblock {\em Psychological bulletin}, 138(6):1085.

\bibitem[Guertler et~al., 2025]{guertler2025textarena}
Guertler, L., Cheng, B., Yu, S., Liu, B., Choshen, L., and Tan, C. (2025).
\newblock Textarena.
\newblock {\em arXiv preprint arXiv:2504.11442}.

\bibitem[Ha and Schmidhuber, 2018]{ha2018world}
Ha, D. and Schmidhuber, J. (2018).
\newblock World models.
\newblock {\em arXiv preprint arXiv:1803.10122}.

\bibitem[Hao et~al., 2023]{hao2023reasoning}
Hao, S., Gu, Y., Ma, H., Hong, J.~J., Wang, Z., Wang, D.~Z., and Hu, Z. (2023).
\newblock Reasoning with language model is planning with world model.
\newblock {\em arXiv preprint arXiv:2305.14992}.

\bibitem[Henderson et~al., 2010]{henderson2010structure}
Henderson, L., Goodman, N.~D., Tenenbaum, J.~B., and Woodward, J.~F. (2010).
\newblock The structure and dynamics of scientific theories: A hierarchical bayesian perspective.
\newblock {\em Philosophy of Science}, 77(2):172--200.

\bibitem[Johnson-Laird, 1983]{johnson1983mental}
Johnson-Laird, P.~N. (1983).
\newblock {\em Mental models: Towards a cognitive science of language, inference, and consciousness}.
\newblock Number~6. Harvard University Press.

\bibitem[Kaelbling et~al., 1998]{kaelbling1998planning}
Kaelbling, L.~P., Littman, M.~L., and Cassandra, A.~R. (1998).
\newblock Planning and acting in partially observable stochastic domains.
\newblock {\em Artificial intelligence}, 101(1-2):99--134.

\bibitem[Kaiser et~al., 2019]{kaiser2019model}
Kaiser, L., Babaeizadeh, M., Milos, P., Osinski, B., Campbell, R.~H., Czechowski, K., Erhan, D., Finn, C., Kozakowski, P., Levine, S., et~al. (2019).
\newblock Model-based reinforcement learning for atari.
\newblock {\em arXiv preprint arXiv:1903.00374}.

\bibitem[Lake et~al., 2017]{lake2017building}
Lake, B.~M., Ullman, T.~D., Tenenbaum, J.~B., and Gershman, S.~J. (2017).
\newblock Building machines that learn and think like people.
\newblock {\em Behavioral and brain sciences}, 40:e253.

\bibitem[LeCun, 2022]{lecun2022path}
LeCun, Y. (2022).
\newblock A path towards autonomous machine intelligence version 0.9. 2, 2022-06-27.
\newblock {\em Open Review}, 62(1):1--62.

\bibitem[Li et~al., 2023]{li2023emergent}
Li, K., Hopkins, A.~K., Bau, D., Vi{\'e}gas, F., Pfister, H., and Wattenberg, M. (2023).
\newblock Emergent world representations: Exploring a sequence model trained on a synthetic task.
\newblock {\em ICLR}.

\bibitem[Liu and Xu, 2024]{liu2024violations}
Liu, R. and Xu, F. (2024).
\newblock Violations of core object principles change adults’ behaviors in maze games.
\newblock In {\em Proceedings of the Annual Meeting of the Cognitive Science Society}, volume~46.

\bibitem[Moerland et~al., 2023]{moerland2023model}
Moerland, T.~M., Broekens, J., Plaat, A., Jonker, C.~M., et~al. (2023).
\newblock Model-based reinforcement learning: A survey.
\newblock {\em Foundations and Trends{\textregistered} in Machine Learning}, 16(1):1--118.

\bibitem[Newell, 1955]{newell1955chess}
Newell, A. (1955).
\newblock The chess machine: an example of dealing with a complex task by adaptation.
\newblock In {\em Proceedings of the March 1-3, 1955, western joint computer conference}, pages 101--108.

\bibitem[Newell et~al., 1972]{newell1972human}
Newell, A., Simon, H.~A., et~al. (1972).
\newblock {\em Human problem solving}, volume 104.
\newblock Prentice-hall Englewood Cliffs, NJ.

\bibitem[Perez-Liebana et~al., 2019]{perez2019general}
Perez-Liebana, D., Liu, J., Khalifa, A., Gaina, R.~D., Togelius, J., and Lucas, S.~M. (2019).
\newblock General video game ai: A multitrack framework for evaluating agents, games, and content generation algorithms.
\newblock {\em IEEE Transactions on Games}, 11(3):195--214.

\bibitem[Plappert et~al., 2018]{plappert2018multi}
Plappert, M., Andrychowicz, M., Ray, A., McGrew, B., Baker, B., Powell, G., Schneider, J., Tobin, J., Chociej, M., Welinder, P., et~al. (2018).
\newblock Multi-goal reinforcement learning: Challenging robotics environments and request for research.
\newblock {\em arXiv preprint arXiv:1802.09464}.

\bibitem[Pourcel et~al., 2023]{pourcel2023aces}
Pourcel, J., Colas, C., Molinaro, G., Oudeyer, P.-Y., and Teodorescu, L. (2023).
\newblock Aces: Generating diverse programming puzzles with with autotelic generative models.
\newblock {\em arXiv preprint arXiv:2310.10692}.

\bibitem[Pugh et~al., 2016]{pugh2016quality}
Pugh, J.~K., Soros, L.~B., and Stanley, K.~O. (2016).
\newblock Quality diversity: A new frontier for evolutionary computation.
\newblock {\em Frontiers in Robotics and AI}, 3:40.

\bibitem[Richens et~al., 2025]{richens2025general}
Richens, J., Abel, D., Bellot, A., and Everitt, T. (2025).
\newblock General agents need world models.
\newblock {\em arXiv preprint arXiv:2506.01622}.

\bibitem[Schaul, 2013]{schaul2013video}
Schaul, T. (2013).
\newblock A video game description language for model-based or interactive learning.
\newblock In {\em 2013 IEEE Conference on Computational Inteligence in Games (CIG)}, pages 1--8. IEEE.

\bibitem[Schulz, 2012]{schulz2012origins}
Schulz, L. (2012).
\newblock The origins of inquiry: Inductive inference and exploration in early childhood.
\newblock {\em Trends in cognitive sciences}, 16(7):382--389.

\bibitem[Shannon, 1950]{shannon1950}
Shannon, C.~E. (1950).
\newblock Xxii. programming a computer for playing chess.
\newblock {\em The London, Edinburgh, and Dublin Philosophical Magazine and Journal of Science}, 41(314):256--275.

\bibitem[Silver et~al., 2016]{silver2016mastering}
Silver, D., Huang, A., Maddison, C.~J., Guez, A., Sifre, L., Van Den~Driessche, G., Schrittwieser, J., Antonoglou, I., Panneershelvam, V., Lanctot, M., et~al. (2016).
\newblock Mastering the game of go with deep neural networks and tree search.
\newblock {\em nature}, 529(7587):484--489.

\bibitem[Spelke and Kinzler, 2007]{spelke2007core}
Spelke, E.~S. and Kinzler, K.~D. (2007).
\newblock Core knowledge.
\newblock {\em Developmental science}, 10(1):89--96.

\bibitem[Tenenbaum et~al., 2006]{tenenbaum2006theory}
Tenenbaum, J.~B., Griffiths, T.~L., and Kemp, C. (2006).
\newblock Theory-based bayesian models of inductive learning and reasoning.
\newblock {\em Trends in cognitive sciences}, 10(7):309--318.

\bibitem[Tenenbaum et~al., 2011]{tenenbaum2011grow}
Tenenbaum, J.~B., Kemp, C., Griffiths, T.~L., and Goodman, N.~D. (2011).
\newblock How to grow a mind: Statistics, structure, and abstraction.
\newblock {\em science}, 331(6022):1279--1285.

\bibitem[Todd et~al., 2024]{todd2024gavel}
Todd, G., Padula, A.~G., Stephenson, M., Piette, {\'E}., Soemers, D.~J., and Togelius, J. (2024).
\newblock Gavel: Generating games via evolution and language models.
\newblock {\em Advances in Neural Information Processing Systems}, 37:110723--110745.

\bibitem[Tsividis et~al., 2021]{tsividis2021human}
Tsividis, P.~A., Loula, J., Burga, J., Foss, N., Campero, A., Pouncy, T., Gershman, S.~J., and Tenenbaum, J.~B. (2021).
\newblock Human-level reinforcement learning through theory-based modeling, exploration, and planning.
\newblock {\em arXiv preprint arXiv:2107.12544}.

\bibitem[Ullman and Tenenbaum, 2020]{ullman2020bayesian}
Ullman, T.~D. and Tenenbaum, J.~B. (2020).
\newblock Bayesian models of conceptual development: Learning as building models of the world.
\newblock {\em Annual Review of Developmental Psychology}, 2(1):533--558.

\bibitem[Vafa et~al., 2024]{vafa2024evaluating}
Vafa, K., Chen, J., Rambachan, A., Kleinberg, J., and Mullainathan, S. (2024).
\newblock Evaluating the world model implicit in a generative model.
\newblock {\em Advances in Neural Information Processing Systems}, 37:26941--26975.

\bibitem[Van~Hasselt et~al., 2016]{van2016deep}
Van~Hasselt, H., Guez, A., and Silver, D. (2016).
\newblock Deep reinforcement learning with double q-learning.
\newblock In {\em Proceedings of the AAAI conference on artificial intelligence}, volume~30.

\bibitem[van Opheusden et~al., 2023]{van2023expertise}
van Opheusden, B., Kuperwajs, I., Galbiati, G., Bnaya, Z., Li, Y., and Ma, W.~J. (2023).
\newblock Expertise increases planning depth in human gameplay.
\newblock {\em Nature}, pages 1--6.

\bibitem[V{\'e}lez et~al., 2024]{velez2024rise}
V{\'e}lez, N., Wu, C.~M., Gershman, S.~J., and Schulz, E. (2024).
\newblock The rise and fall of technological development in virtual communities.

\bibitem[Verma et~al., 2025]{verma2025measuring}
Verma, V., Huang, D., Chen, W., Klein, D., and Tomlin, N. (2025).
\newblock Measuring general intelligence with generated games.
\newblock {\em arXiv preprint arXiv:2505.07215}.

\bibitem[Vinyals et~al., 2019]{vinyals2019grandmaster}
Vinyals, O., Babuschkin, I., Czarnecki, W.~M., Mathieu, M., Dudzik, A., Chung, J., Choi, D.~H., Powell, R., Ewalds, T., Georgiev, P., et~al. (2019).
\newblock Grandmaster level in starcraft ii using multi-agent reinforcement learning.
\newblock {\em nature}, 575(7782):350--354.

\bibitem[Wang et~al., 2023]{wang2023scientific}
Wang, H., Fu, T., Du, Y., Gao, W., Huang, K., Liu, Z., Chandak, P., Liu, S., Van~Katwyk, P., Deac, A., et~al. (2023).
\newblock Scientific discovery in the age of artificial intelligence.
\newblock {\em Nature}, 620(7972):47--60.

\bibitem[Yannakakis and Togelius, 2018]{yannakakis2018artificial}
Yannakakis, G.~N. and Togelius, J. (2018).
\newblock {\em Artificial intelligence and games}, volume~2.
\newblock Springer.

\bibitem[Youn et~al., 2015]{youn2015invention}
Youn, H., Strumsky, D., Bettencourt, L.~M., and Lobo, J. (2015).
\newblock Invention as a combinatorial process: evidence from us patents.
\newblock {\em Journal of the Royal Society interface}, 12(106):20150272.

\bibitem[Yu et~al., 2020]{yu2020meta}
Yu, T., Quillen, D., He, Z., Julian, R., Hausman, K., Finn, C., and Levine, S. (2020).
\newblock Meta-world: A benchmark and evaluation for multi-task and meta reinforcement learning.
\newblock In {\em Conference on robot learning}, pages 1094--1100. PMLR.

\bibitem[Zhang et~al., 2024]{zhang2024people}
Zhang, C.~E., Collins, K.~M., Wong, L., Barba, M., Weller, A., and Tenenbaum, J.~B. (2024).
\newblock People use fast, goal-directed simulation to reason about novel games.
\newblock {\em arXiv preprint arXiv:2407.14095}.

\bibitem[Zhao et~al., 2024]{zhao2024rational}
Zhao, B., V{\'e}lez, N., and Griffiths, T. (2024).
\newblock A rational model of innovation by recombination.
\newblock In {\em Proceedings of the Annual Meeting of the Cognitive Science Society}, volume~46.

\bibitem[Zhu et~al., 2024]{zhu2024sora}
Zhu, Z., Wang, X., Zhao, W., Min, C., Deng, N., Dou, M., Wang, Y., Shi, B., Wang, K., Zhang, C., et~al. (2024).
\newblock Is sora a world simulator? a comprehensive survey on general world models and beyond.
\newblock {\em arXiv preprint arXiv:2405.03520}.

\end{thebibliography}
\bibliographystyle{apalike}

\end{document}